\documentclass[a4paper,fleqn]{cas-dc}

\usepackage[authoryear,longnamesfirst]{natbib}

\usepackage{algorithm}
\usepackage{algorithmic}

\def\tsc#1{\csdef{#1}{\textsc{\lowercase{#1}}\xspace}}
\tsc{WGM}
\tsc{QE}

\begin{document}
\let\WriteBookmarks\relax
\def\floatpagepagefraction{1}
\def\textpagefraction{.001}

\shorttitle{}    

\shortauthors{}  

\title [mode = title]{SCGNN: Semantic Consistency enhanced Graph Neural Network Guided by Granular-ball Computing}  



%

\author[1]{Genhao Tian}[orcid=0009-0006-0249-8983]



\ead{241110702108@stu.just.edu.cn}



\affiliation[1]{organization={JiangSu University of Science and Technoloy},
            city={Zhenjiang},
            state={Jiangsu},
            country={China}}

\affiliation[2]{organization={Chongqing University of Posts and Telecommunications},
            city={Chongqing},
            country={China}}

\affiliation[3]{organization={Zunyi Normal University},
            city={Zunyi},
            state={Guizhou},
            country={China}}

\author[1]{Taihua Xu}
\cormark[1]

\ead{xutaihua2019@just.edu.cn}




\cortext[1]{Corresponding author}

\author[2]{Shuyin Xia}
\ead{xiasy@cqupt.edu.cn}

\author[2]{Qinghua Zhang}
\ead{zhangqh@cqupt.edu.cn}

\author[3]{Jie Yang}
\ead{2012009@zync.edu.cn}

\author[1]{Jianjun Chen}
\ead{jianjunchen@just.edu.cn}



\begin{abstract}
Capturing semantic consistency among nodes is crucial for effective graph representation learning. Existing approaches typically rely on $k$-nearest neighbors ($k$NN) or other node-level full search algorithms (FSA) to mine semantic relationships via exhaustive pairwise similarity computation, which suffer from high computational complexity and rigid neighbor selection, limiting scalability and introducing noisy connections. In this paper, we propose the Semantic Consistency enhanced Graph Neural Network (SCGNN), a novel plug-and-play framework that leverages granular-ball computing (GBC) to efficiently capture semantic consistency in a scalable manner. Unlike node-level FSA methods, SCGNN models group-level semantic structure by adaptively partitioning nodes into granular balls, significantly reducing computational cost while improving robustness to noise. To effectively utilize the discovered group-level semantic consistency, we design a dual enhancement strategy. Specifically, (1) a structure enhancement module constructs an anchor-based graph structure, where each anchor is a virtual node representing the group-level semantic carried by a granular ball, then injecting group-level semantic information into the graph structure; and (2) a supervision enhancement module performs label consistency checking (LCC) by combining GBC predictions with model-generated pseudo-labels, thereby producing more reliable supervision signals. SCGNN is compatible with various GNN backbones. During the forward propagation of SCGNN, the vanilla graph and the augment graph are jointly encoded, and their predictions are fused; during the backpropagation, the supervision enhancement module provides enhanced supervision signals to guide parameter updates. Extensive experiments on every benchmarks demonstrate that SCGNN consistently improves performance while maintaining high computational efficiency and scalability.

\end{abstract}





\maketitle

\section{Introduction}
\label{intro}

Graph-structured data is widely encountered in real-world applications, and graph neural networks (GNNs) \cite{gcn,gat,appnp} have become a dominant paradigm for learning representations from such data. Depending on the task granularity, GNNs have been successfully applied to node-level tasks \cite{gcn}, edge-level tasks \cite{edgeclass1,edgeclass2}, and graph-level tasks \cite{gin,graphclass1,graphclass2}. Despite their differences, these approaches generally rely on learning informative node embeddings through message passing over graph structure.

However, such structure-driven message passing primarily exploits the given graph structure and may fail to fully capture rich semantic relationships embedded in node features. Therefore, relying solely on the observed graph structure is insufficient, and mining semantic information from node features has emerged as an important research direction in graph representation learning \cite{TKDE24, PR24, geomgcn_webkb, comfy,violin,wang-eswa25}.

The semantic mining direction mainly revolve around two key aspects: (1) the extraction of semantic consistency information, i.e., how to efficiently mine high-quality semantic consistency from features, and (2) the utilization of such information, i.e., how to properly incorporate the extracted semantic consistency into graph neural networks (GNNs) to enhance representation learning.

Node-level full search algorithms (FSA) are widely used for semantic mining \cite{kNN-GNN-ESWA24, ledf, ooc-icml24}. These methods operate at a very fine granularity, where the semantic consistency (feature similarity) between every pair of samples is computed. Consequently, in the utilization phase, they naturally modify the graph structure based on pairwise similarity. Specifically, for each node, the $top$K most similar candidates are selected as its neighbors (commonly referred to as $k$NN or $top$K), or neighbors are filtered based on a predefined similarity threshold.

In theory, FSA can extract the highest-quality semantic consistency information. However, they suffer from notable limitations in both extraction and utilization. On the extraction side, their quadratic complexity leads to prohibitive computational costs as the number of samples increases \cite{iclr24, kNN-GNN-ESWA24}. On the utilization side, $k$NN or $top$K methods rely on a predefined k, which introduces hyperparameter sensitivity and $k$-value noise \cite{knoise-aaai23, wang-eswa25}. Similarity threshold based methods provide better adaptivity at the node level, but determining an appropriate threshold is often challenging.

To reduce the computational costs of FSA in semantic consistency mining, we introduce a highly efficient approach—Granular Ball Computing (GBC) \cite{gbc-is-xia19,gbc-tnnls-xia22,gbc-tnnls-xia25,gbc-tnnls-cheng23}. GBC can alleviate the FSA scalability bottleneck while also mitigate the challenges of hyperparameter selection and $k$-value noise in the utilization phase.

Unlike node-level FSA, which performs one-shot global computation, GBC adopts an iterative process that progressively splits granular balls (GBs) until all of them satisfy a predefined stopping criterion \cite{gbc-is-xia19}. The most significant advantage of GBC lies in its scalability. As shown in Table \ref{tab.scale}, we compare the overhead of GBC and $k$NN across datasets of different scales. Even with up to 90 GB of RAM, $k$NN fails to process the million-scale ogbn-arxiv dataset, whereas GBC can efficiently handle the ten-million-scale ogbn-products dataset with low memory overhead even lower than the overhead of $k$NN on the much smaller arxiv2023 dataset with only 46k samples.
Notably, GBC is slightly slower than $k$NN only on Amazon-ratings dataset, mainly because $k$NN benefits from highly optimized matrix operations, whereas GBC incurs additional fixed overhead from its recursive splitting. As the scale increases (e.g., arxiv2023 and beyond), this overhead becomes negligible which makes GBC quickly outperforms $k$NN.

\begin{table}
\caption{Comparison of time and memory overhead of GBC and $k$NN on datasets of different scales. We have a total of 92160MB (90GB) of RAM memory.}
\label{tab.scale}
\begin{tabular*}{\tblwidth}{@{} LLLLL @{}}
\toprule
\multirow{2}{*}{Datasets(scale)}&\multicolumn{2}{L}{Time(s)}&\multicolumn{2}{L}{Memory(MB)} \\
\cmidrule{2-3}
\cmidrule{4-5}
&GBC&$k$NN&GBC&$k$NN \\
\midrule
Amazon-ratings(24k)&20&12&227&3995\\
arxiv2023(46k)&30&33&379&10538\\
ogbn-arxiv(169k)&106&N/A&597&$>$92160\\
ogbn-products(2449k)&1035&N/A&7109&$\gg$ 92160\\
\bottomrule
\end{tabular*}
\end{table}

The substantial gap in computational overhead between $k$NN and GBC demonstrates the superior scalability of GBC. However, the group-level consistency discovered by GBC cannot be directly used to select neighbors based on pairwise similarity, as in FSA. Therefore, effectively utilizing group-level consistency is the key to integrating GBC into GNNs.

Based on the scalability of GBC and the utilization of its group-level semantic consistency, we propose the Semantic Consistency enhanced Graph Neural Network (SCGNN). First, we apply GBC to extract group-level semantic consistency. Then, for semantic utilization, we design a dual enhancement strategy consisting of structure enhancement and supervision enhancement.

Specifically, for structure enhancement, we design an anchor-based two-level graph structure construction strategy, where each anchor is a virtual node representing the group-level semantics carried by a granular ball (GB). The two-level graph structure consists of (1) projection edges connecting each anchor to the nodes within its corresponding granular ball, and (2) bridging edges linking anchors that share similar semantics, which inject group-level semantic consistency into the graph structure and enables GNNs to exploit semantic consistency via spectral information. 
For supervision enhancement, we perform GB-based label prediction, and use the proposed Label Consistency Check (LCC) which combines GB label with the pseudo-label \cite{pseudo-label} from model predictions, to produce LCC labels for additional supervision.  

In the overall architecture of SCGNN, during the forward propagation, we adopt a parallel prediction and prediction fusion scheme based on both the vanilla graph and the augment graph. During the backpropagation, the model is jointly supervised by train labels and LCC labels. Furthermore, the model is designed in a plug-and-play manner, enabling SCGNN to be compatible with various GNN backbones.

The main contributions are summarized as follows:
\begin{enumerate}

\item \textbf{A new perspective for semantic consistency mining.}
We introduce granular-ball computing (GBC) into graph representation learning, providing a scalable alternative to node-level FSA for capturing semantic consistency.

\item \textbf{A dual enhancement strategy for utilizing group-level semantic consistency.}
To effectively exploit the group-level semantic consistency discovered by GBC, we design a dual enhancement strategy: 
(1) a \emph{structure enhancement} module that constructs an anchor-based two-level graph structure to inject semantic consistency into the graph structure, and (2) a \emph{supervision enhancement} module that combines GB predictions with model-generated pseudo-labels via a Label Consistency Check (LCC).

\item \textbf{A plug-and-play GNN framework.}
SCGNN is designed in a plug-and-play manner which enable it to be compatible with various GNN backbones. 

\item \textbf{Extensive experimental validation.}
Extensive experiments on multiple benchmark datasets show that SCGNN consistently outperforms state-of-the-art methods while maintaining high computational efficiency and scalability.
\end{enumerate}

The source code is publicly available at: \url{https://github.com/GCNJust/SCGNN}.

\section{Related Work}
\label{related}

\subsection{Semantic consistency research of GNN}
Several studies attempt to exploit semantic consistency in node features to augment or refine graph structure. For instance, AM-GCN\cite{amgcn}: Adaptive Multi-channel Graph Convolutional Networks constructs a feature similarity graph by computing k-nearest neighbors in the vanilla feature space, thereby introducing additional edges beyond the vanilla graph structure to capture potential semantic relations between nodes. Geom-GCN\cite{geomgcn_webkb}: Geometric Graph Convolutional Networks embeds nodes into a latent geometric space and reorganizes neighborhoods based on geometric proximity, enabling the model to capture semantic relationships that may not align with the vanilla graph connectivity. IDGL\cite{idgl}: Iterative Deep Graph Learning proposes to iteratively learn an optimized graph structure from node features by refining edge weights based on feature similarity, allowing the model to better capture intrinsic semantic relationships and alleviate the limitations of the vanilla graph structure.

\subsection{Granular-ball computing}
Granular-Ball Computing (GBC) was first introduced as a data representation and learning framework grounded in Granular Computing \cite{gbc-is-xia19}. Instead of treating individual samples as the basic computational units, GBC represents data using granular balls, each of which corresponds to a locally semantically consistent region. A granular ball summarizes its contained samples through statistical properties such as the center, radius, and label distribution, enabling learning at a higher level of granularity. This paradigm effectively preserves structural information while improving computational efficiency and robustness to noise.

Subsequent works have focused on improving the construction and scalability of granular balls. Xia et al. \cite{gbc-tnnls-xia22, gbc-tnnls-xia25} proposed adaptive granular-ball generation methods that iteratively split the data space based on criteria such as purity and data distribution, enabling flexible and efficient granulation for complex data. These methods significantly enhance the stability and adaptability of GBC in practical scenarios.

To further extend GBC to large-scale unsupervised settings, Cheng et al. \cite{gbc-tnnls-cheng23} introduced a size-constrained granular-ball construction strategy, where the number of samples within each ball is bounded (e.g., $O(\sqrt{n})$), removing the reliance on label information. Based on this representation, they combined GBC with density peaks clustering, performing density estimation and clustering at the granular-ball level to substantially reduce computational cost on large-scale datasets.

\section{Methodology}
\label{methodology}

In this section, we first introduce the notations and functions in section \ref{sec.pre}, then we will introduce our method starting from the GBC semantic consistency estimator in section \ref{sec.sce}, finally, the data enhancement strategy and the semantic consistency graph neural network (SCGNN) will be introduced in section \ref{sec.aug_data} and  \ref{sec.scgnn}.

\subsection{Preliminary}
\label{sec.pre}
In this section, we give all functions and notations we use in the equations of this paper, presented as Table \ref{tab.notation}.

\begin{table}
    \caption{Notations and descriptions.}
    \begin{tabular*}{\tblwidth}{@{} LL @{} }
        \toprule
        Notatation& Description \\
        \midrule
         $x$& a node\\
         $\mathbf{G}$& the vanilla graph\\
         $\mathbf{G^{aug}}$& the augment graph\\
         $\mathbf{L}$& the label vector\\
         $\mathbf{F}$& the features matrix\\
         $\mathbf{GBs}$& the generated granular balls set\\
         $\mathbf{GB_k}$& the $k-$th granular ball\\
         $l(\mathbf{GB} / x)$& the label of $\mathbf{GB}$ or node $x$\\
         $p(\mathbf{GB})$& the purity of a granular ball\\
         $\mathbf{GNN(\cdot)}$& an classic GNN model\\
         $\mathbf{GBC(\cdot)}$& GBC semantic consistency estimator\\
         \bottomrule
    \end{tabular*}
    \label{tab.notation}
\end{table}

We give the functions and notations usually used in GNN fields first. The origin graph $\textbf{G} = (\textbf{V}, \textbf{E})$, $\mathbf{V}$ denotes the node set and $\textbf{E}$ denotes the edge set, and $\textbf{G}^{\textbf{aug}} = (\textbf{V}^{\textbf{aug}}, \textbf{E}^{\textbf{aug}})$ is the augment graph, $\mathbf{G \subset G^{aug}}$. All features are represented by a matrix $\textbf{F} \in \mathbb{R}^{n\times d}$ where $n = |\textbf{V}|$ is the number of nodes and $d$ is the number of features. Each node has only one label represented as a label vector $\textbf{L}$. 

$\mathbf{GNN(\cdot)}$ denotes any GNN model, the predictions of GNNs is $\textbf{P}\in\mathbb{R}^{n\times c}$.
$\mathbf{GBC(\cdot)}$ is the semantic consistency estimator guided by the granular ball computing, and the predictions of the estimator is $\textbf{P}^{\textbf{gbc}}\in \mathbb{R}^{n'\times c}$.

\subsection{GBC semantic consistency estimator}
\label{sec.sce}
In this section, we will introduce the granular-ball computing (GBC) we used in this paper, and next build on the GBC introduce the GBC semantic consistency estimator.

We feed all samples into GBC in a transductive manner, including features of all samples and labels of the training samples, to perform granular-ball computing,
before formulating the GBC, we need to introduce an important threshold index in granular-ball computing which is purity and it is computed as below:
\begin{equation}
    p(\mathbf{GB})=\frac{\underset{y_j\in \mathbf{GB}_k}{max}   \sum\limits_{x_i\in\mathbf{GB}} \amalg (l(x_i)=y_j)}{|\mathbf{GB}|}.
    \label{eq.purity}
\end{equation}
In particular, if there is no labeled node inside the $\mathbf{GB}$, the $p(\mathbf{GB})$ will be directly set to $1$ to terminate the continued division of it.

For intuitive and rigorous expression, we will introducing the semi-supervised GBC we adapted and used via algorithm \ref{alg1}. The $k$-means algorithm we use is based on the euclidean distance.
\begin{algorithm}[!ht]
\caption{GBC}
\label{alg1}
\textbf{Input:} Vanilla features matrix $\mathbf{F}$, Train nodes label $\mathbf{L^{train}}$.\\
\textbf{Parameter:} the purity threshold $t=0.8$. \\
\text{\textbf{Output:} $\mathbf{GBs}$.}
\begin{algorithmic}
\STATE Initialization an queue with one element $\mathbf{GBs} = [\{f_{1}, f_{2},...,f_{n} \}]$, $f\in \textbf{F}$.\\
\textbf{Do}\\
\hspace*{1em} Let $num$ equal to the length of the queue $\mathbf{GBs}$, i.e., $num = |\mathbf{GBs}|$.\\
\hspace*{1em} \textbf{For} each $\mathbf{GB}$ in $\mathbf{GBs}$\\
\hspace*{2em} Let $p = p(\mathbf{GB})$, by Eq. \ref{eq.purity}.\\
\hspace*{2em} \textbf{IF} $p\le t$\\
\hspace*{3em} Let $k$ equal to the classes number of all labeled nodes in current $\mathbf{GB}$.\\
\hspace*{3em} Dequeue $\mathbf{GB}$ from $\mathbf{GBs}$ queue.\\
\hspace*{3em} Employ $k$-means algorithm to divide one $\mathbf{GB}$ into $k$ little $\mathbf{sub\_GBs}$.\\
\hspace*{3em} Enqueue $\mathbf{sub\_GBs}$ to $\mathbf{GBs}$.\\
\hspace*{2em} \textbf{END}\\
\hspace*{1em} \textbf{END}\\
\textbf{While} The length of the queue no longer changes, i.e., $num$ equal to $|\mathbf{GBs}|$

\textbf{Do}\\
\hspace*{1em} Let $num$ equal to the length of the queue $\mathbf{GBs}$, i.e., $num = |\mathbf{GBs}|$.\\
\hspace*{1em} \textbf{For} each $\mathbf{GB}$ in $\mathbf{GBs}$\\
\hspace*{2em} \textbf{IF} $|GB|\ge \sqrt{n}$\\
\hspace*{3em} Dequeue $\mathbf{GB}$ from $GBs$ queue.\\
\hspace*{3em} Employ $2$-means algorithm to divide one $\mathbf{GB}$ into $2$ little $\mathbf{sub\_GBs}$.\\
\hspace*{3em} Enqueue $\mathbf{sub\_GBs}$ to $\mathbf{GBs}$.\\
\hspace*{2em} \textbf{END}\\
\hspace*{1em} \textbf{END}\\
\textbf{While} The length of the queue no longer changes, i.e., $num$ equal to $|\mathbf{GBs}|$
\end{algorithmic}
\end{algorithm}

We combine supervised granular-ball computing \cite{gbc-is-xia19} and unsupervised granular-ball computing \cite{gbc-tnnls-cheng23}, the advantage of this is that unsupervised method will split the oversized $GB$ produced by the supervised method to ensure that the size of the $GB$ remains uniform to a certain extent.

$\mathbf{GB}$ may have the label, it is computed as fallow:
\begin{equation}
    l(\mathbf{GB})=\underset{y_j\in \mathbf{GB}_k}{argmax}   \sum\limits_{x_i\in\mathbf{GB}} \amalg (l(x_i)=y_j).
    \label{eq.plabel}
\end{equation}

Traditional Granular Ball Computing (GBC) adopts an inductive paradigm, where only training samples are used to construct granular balls. Each ball is explicitly modeled as $\mathbf{GB} = (c, r, l(\mathbf{GB}))$, $c$ is the center of $\mathbf{GB}$, $r$ is the radius of $\mathbf{GB}$, and test samples are subsequently classified based on their nearest ball.

\begin{figure}
    \centering
    \includegraphics[width=0.9\columnwidth]{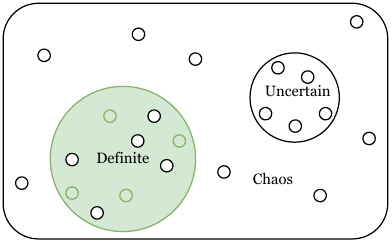}
    \caption{The illustration of three kinds of semantic domain, green circle is definite domain, white circle is uncertain domain and out of circle is chaos domain.}
    \label{fig.three}
\end{figure}

In contrast, our objective is not to build an explicit granular ball classifier, but to exploit the scalability and robustness of GBC for estimating local semantic consistency. Therefore, we adopt a transductive GBC strategy, where both training and unlabeled samples participate in the granular formation process, and the transductive strategy has another strength, it can directly use the coverage characteristics of GBC to measure the uncertainty of local semantic consistency. And, owing to this intrinsic properties, all nodes are ultimately assigned to one of three semantic domains: the semantic definite domain: covered by GB; the semantic uncertain domain: covered by GB but without label; and the semantic chaos domain: not covered by any GB; as illustration of Figure \ref{fig.three}. This characteristic naturally provides a principled way to quantify the uncertainty of local semantic consistency estimation.

Based on these three semantic domains, we can intuitively and reasonably divide all nodes into two categories: nodes in the semantic definite domain we called it UG(under GB) nodes an the others in the semantic uncertain and semantic chaos domain we called it OOG(out of GB) nodes. 

UG nodes are more likely to share consistent local semantics with the labeled samples inside the same ball. In contrast, OOG nodes typically correspond to semantically ambiguous or feature-complex instances that are difficult to separate based solely on feature similarity, these instances tend to exhibit higher noise levels. Due to the inherent robustness of GBC, such noisy or uncertain samples can be effectively identified through the coverage mechanism.

Based on the motivations above, the GBC semantic consistency estimator will be introduced. We provide a formal description of the GBC semantic consistency estimator by integrating the semi-supervised granular-ball computing method introduced above. The GBC (algorithm \ref{alg1}) denoted as $\mathbf{GBC(\cdot)}$ receive the features of samples $\mathbf{F}$ and the supervision information $\mathbf{L^{train}}$ and last return the $\mathbf{GBs}$ after computing.

\begin{equation}
    \label{eq.gbc}
    \mathbf{GBs} = \mathbf{GBC(F, L^{train})}.
\end{equation}

According the semantic consistency, the label $l(\mathbf{GB})$ i.e., the most common supervised label in each $\mathbf{GB}$ can be treated as the label of all non-label nodes within this $\mathbf{GB}$, simply put, each $\mathbf{GB}$ is treated as a local label domain.

\begin{equation}
    \label{eq.lse}
    \mathbf{P^{gbc}_{j}} = l(\mathbf{GB}), j\in \mathbf{GB}.
\end{equation}

$\mathbf{j}$ is the sample under $\mathbf{GB}$, and $\mathbf{P^{gbc}_j}\in \mathbb{R}^{n' \times c}$, because of the highly uncertainty of OOG nodes, we gave up label prediction for OOG nodes, so $n'$ is less than the total sample number $n$, this will make the predictions of GBC much more reliable.

\subsection{Semantic consistency enhancement strategy}
\label{sec.aug_data}

In this section we will introduce our data enhancement strategy, it contains two parts, one is the structure enhancement and the other is the supervision enhancement, these two will be introduced in turn.

\begin{figure*}
    \centering
    \includegraphics[width=\textwidth]{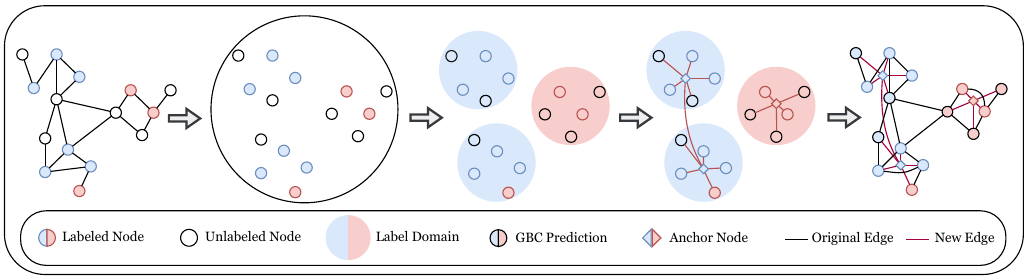}
    \caption{The process of the structure enhancement, four arrows represent GB initialization, GBC predicting, new edges generation and output augment graph.}
    \label{fig.aug}
\end{figure*}

\subsubsection{Structure enhancement}
The structure enhancement constructs high-quality information propagation pathways based on local semantic consistency, enabling cross-hop connections that capture long-range dependencies beyond the reach of conventional GNNs. The process is illustrated in Figure \ref{fig.aug}. 

Granular-ball computing models feature semantics at the group level. Therefore, explicit pairwise neighbor relations between nodes are not directly available. 

So, we design a novel two-level graph structure construction mechanism that converts granular-ball distributions into graph connections. The construction of augment graph consists of two main steps:

\textbf{First}, selecting the UG training nodes belonging to the label of corresponding GB and compute their mean feature as an anchor representing these UG nodes, that is:
\begin{equation}
    \mathbf{F_i^{anchor}} = \frac{1}{n}\sum_{j=1}^nf_j,j\in \mathbf{GB}_i,\mathbf{V_j}\in \mathbf{V^{train}}.
\end{equation}
And the label of each anchor is assigned to be the label of corresponding GB:
\begin{equation}
    \mathbf{L^{anchor}_k}=l(\mathbf{GB}).
\end{equation}

\textbf{Second}, based on the anchors generated above, we construct the high-quality pathways via a two level graph structure, the first level is the projection edges $\mathbf{E^{project}}$ which is responsible for communicating to the vanilla nodes within each $\textbf{GB}$ from the anchor, and the second level is the bridging edges $\mathbf{E^{bridge}}$ which is responsible for connecting all anchors which are belonging to the same label (on large scale dataset, for controlling the number of bridging edges, the full connection could be replaced by random $k$ edges connection). 

\begin{figure}
    \centering
    \includegraphics[width=0.9\columnwidth]{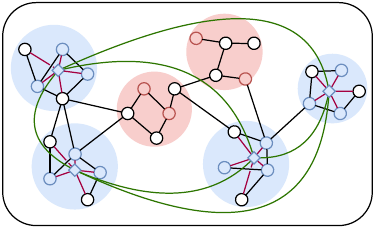}
    \caption{Anchor-based graph structure illustration. For clarity, only the edges associated with blue-labeled nodes is shown. Green edges are bridging edges between anchor nodes, red edges are projection edges.  }
    \label{fig.bridge}
\end{figure}

In this way, vanilla nodes covered by the label domain will get the long distance dependencies in several steps through the high-quality pathways, the illustration is in the Figure \ref{fig.bridge}.

Based on additional anchors and high-quality pathways, the augment graph $\mathbf{G^{aug}}$ can be constructed as fallow:
\begin{equation}
    \mathbf{G^{aug}} = (\mathbf{V^{aug}}, \mathbf{E^{aug}}),
\end{equation}
which $\mathbf{V^{aug}}$ is the corresponding node samples of the $\mathbf{F^{aug}}$ and $\mathbf{F^{aug} = [F|F^{anchor}]}$, $[\cdot|\cdot]$ represent the concatenation. Furthermore, we do not take the radical approach of pruning the vanilla graph structure based on semantic consistency, in other words, the augment graph $\mathbf{G^{aug}}$ contains the entire vanilla graph $\mathbf{G}$, so the $\mathbf{E^{aug} = [E|E^{project}|E^{birdge}]}$, $[\cdot|\cdot|\cdot]$ represent the concatenation.

\subsubsection{Supervision enhancement}
\label{supervision}
The supervision enhancement generates a kind of high-reliability pseudo-label which can provide more supervision information beyond the training label.

Given one GNNs model $\mathbf{GNN_\theta(\cdot)}$, there will be predictions of the input $\mathbf{G}$, i.e.,
\begin{equation}
    \label{eq.gnn}
    \mathbf{P} = \mathbf{GNN_\theta(G)}, \mathbf{P}\in \mathbb{R}^{n\times c},
\end{equation}
and based on the section of GBC semantic consistency estimator specifically in equation \ref{eq.gbc} and \ref{eq.lse} we got the predictions  $\mathbf{P^{gbc}}\in \mathbb{R}^{n'\times c}$ by the estimator, specially, the number of the predictions of the estimator is $n'$, less than the predictions of the GNN model due to the uncertainty measure. We finally got two predictions from different paradigms, one is GNN paradigm which usually emphasizes the effect of graph structure, and another is granular-ball computing paradigm which emphasizes the feature semantic and also more robust and explainable. Because these two predictions come from different paradigms, we think they can find different patterns in one same data and then complement each other, so based on $\mathbf{P}$ and $\mathbf{P^{gbc}}$, we adopt the label consistency checking (LCC) between these two and then get the result $\mathbf{P^{lcc}}$, that is:

\begin{equation}
    \label{eq.scc}
        \mathbf{P^{lcc}}_{i} = \begin{cases}
     \mathbf{P_i},& \mathbf{P_i} = \mathbf{P_i^{gbc}};\\
     \mathbf{N/A},& \mathbf{P_i} \ne \mathbf{P_i^{gbc}}.
    \end{cases}
\end{equation}

$\mathbf{N/A}$ means an invalid prediction that will be filtered, so the result of LCC $\mathbf{P^{lcc}}\in \mathbb{R}^{n'' \times c}$, the number of $\mathbf{P^{lcc}}$ will be less than the number of $\mathbf{P^{gbc}}$, i.e., $n'' \le n' \le n$. But the noise of $\mathbf{P^{lcc}}$ can be lower.

\newtheorem{proposition}{Proposition}

\begin{proposition}
\label{proposition}
Consider a $c$-class classification problem.
Assume two predictors produce labels with independent symmetric noise rates $R_1$ and $R_2$. We perform element-wise label consistency checking (LCC) and retain only the samples whose two predicted labels are identical. Let $R_3$ denote the noise rate among the retained samples. Then
\begin{equation}
        R_3 = \frac{\frac{1}{c-1}\cdot R_1R_2}{(1-R_1)(1-R_2)+\frac{1}{c-1}\cdot R_1R_2}.\\
\end{equation}

When $R_1, R_2 \ll 1$, the denominator is dominated by $(1-R_1)(1-R_2) \approx 1$, leading to the approximation.
\begin{equation}
    R_3 \approx \frac{R_1R_2}{c-1}.\\
\end{equation}
\end{proposition}

\newproof{pot}{Proof of Proposition\ref{proposition}}

\begin{pot}
Let $V$ denote the event that the two predictions agree (i.e., the sample is retained after LCC).

Let $N$ denote the event that the retained label is noisy (i.e., different from the ground-truth label).

By definition, the noise rate among retained samples is
\begin{equation}
    R_3 = P(N|V).
\end{equation}
Using conditional probability,
\begin{equation}
    R_3 = \frac{P(N \cap V)}{P(V)}.
\end{equation}

A noisy retained label occurs only when both predictors produce the same incorrect label. The probability that both predictors are wrong is $R_1R_2$.
Assuming that incorrect predictions are uniformly distributed over the remaining $c-1$ classes,
the probability that two incorrect predictions coincide is $\frac{1}{c-1}$.
Thus,
\begin{equation}
\label{eq.quadratic_noise}
    P(N \cap V) = \frac{R_1R_2}{c-1}.
\end{equation}
The retained samples consist of correct agreements and coincident errors:
\begin{equation}
    P(V) = (1-R_1)(1-R_2) + \frac{R_1R_2}{c-1}.
\end{equation}
Which completes the proof.

\end{pot}

\newproof{discussion}{Discussion of Proposition\ref{proposition}}

\begin{figure}
    \centering
    \includegraphics[width=0.9\columnwidth]{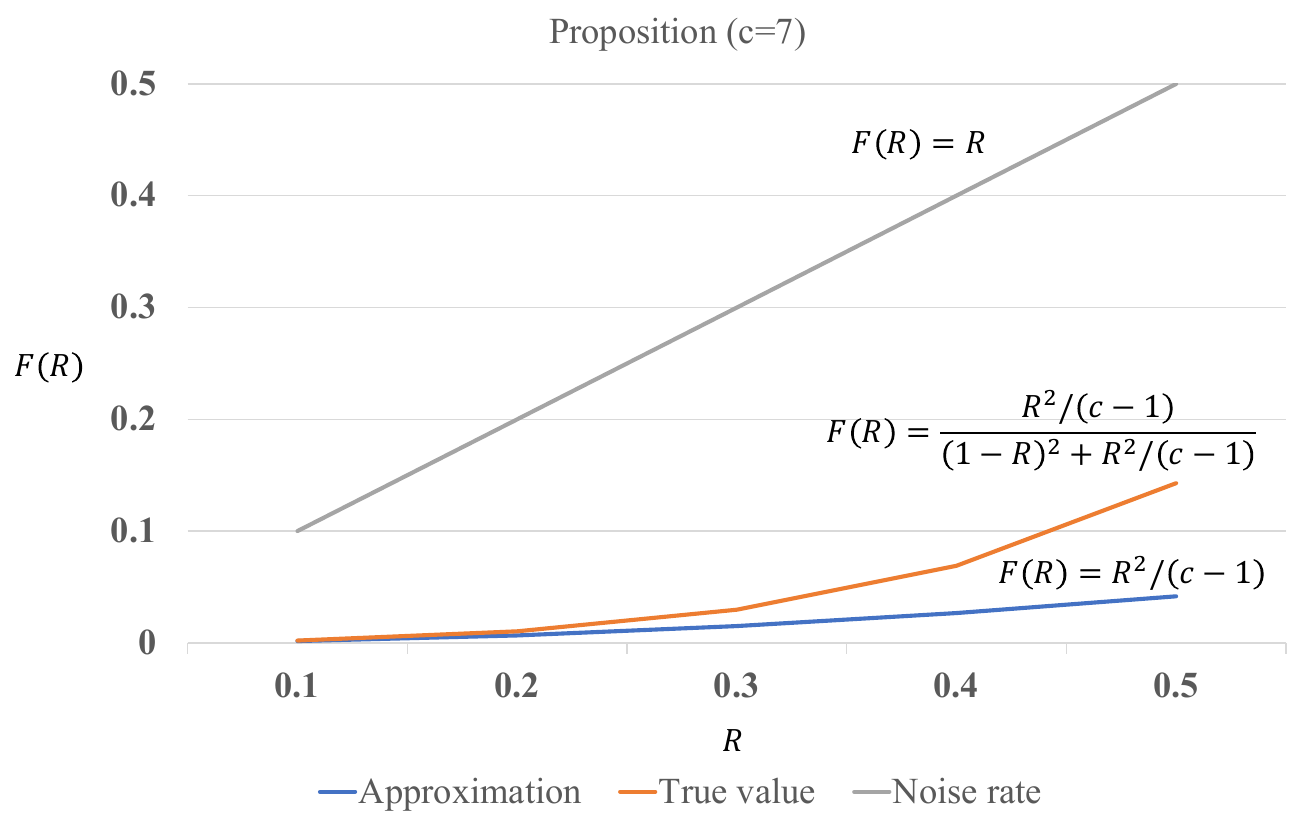}
    \caption{Illustration of the noise reduction effect of LCC with $c=7$ classes.}
    \label{fig.proposition}
\end{figure}

\begin{discussion}
This result shows that LCC reduces the noise rate from a linear dependence on individual error rates to a quadratic dependence.
Specifically, as shown in Eq. \ref{eq.quadratic_noise}, a noisy retained sample occurs only when both predictors make incorrect predictions and their errors coincide, which is significantly less likely than a single prediction error. 
Therefore, LCC significantly reduces the proportion of noisy labels while preserving correct predictions. 
Moreover, the reduction effect becomes more pronounced as the number of classes increases, since the coincidence probability is inversely proportional to $c-1$.
This behavior is further validated by the numerical results shown in Figure \ref{fig.proposition}.

This assumption simplifies analysis, although in practice the predictions may exhibit some correlation.

\end{discussion}

\subsection{Semantic consistency enhanced graph neural network}
\label{sec.scgnn}
In this section, we rename $\mathbf{P^{lcc}}$ to $\mathbf{L^{lcc}}$, because we will treat it as the pseudo label of the samples next. 

\begin{figure}
    \centering
    \includegraphics[width=0.9\columnwidth]{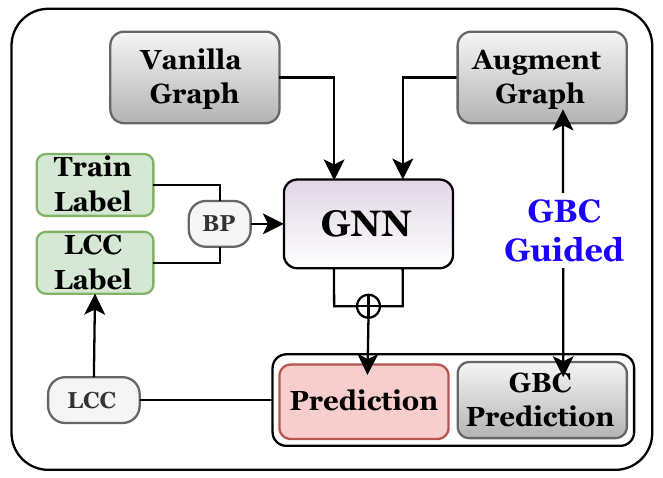}
    \caption{The pipeline of the SCGNN. The augment graph and GBC prediction are guided by granular-ball computing.}
    \label{fig.scgnn}
\end{figure}

Based on the augment graph $\mathbf{G^{aug}}$, anchor label $\mathbf{L^{anchor}}$ , pseudo label $\mathbf{L^{lcc}}$ and the origin data, the semantic consistency enhancement graph neural network (SCGNN) has been proposed, as illustration in Figure \ref{fig.scgnn}.

Starting with forward propagation, SCGNN adopts parallel strategy of vanilla graph $\mathbf{G}$ and augment graph $\mathbf{G^{aug}}$, and share the parameters of neural network, their corresponding predictions are computed as fallow:

\begin{equation}
    \begin{cases}
     \mathbf{P^{aug}}=\mathbf{GNN_\theta(\mathbf{G^{aug}})};\\
     \mathbf{P=\mathbf{GNN_\theta(\mathbf{G})}},
    \end{cases}
    \label{eq.parallel}
\end{equation}

After that, an attention based node-wise weighted sum are utilized for fusing these two kinds of predictions, the fused prediction $\mathbf{P^{fuse}}$ is computed as:
\begin{equation}
    \mathbf{P^{fuse}} = diag(\alpha)\cdot \mathbf{P^{}} + diag(1-\alpha)\cdot \mathbf{P^{aug}},
\end{equation}
$diag(\cdot)$ means diagonalization, and $\mathbf{P^{fuse}}$ is the final prediction of SCGNN, the weights are computed as:
\begin{equation}
\boldsymbol{\alpha} =\frac{exp(\sigma(\mathbf{P}W_{1})W_{2})}{exp\left( \sigma(\mathbf{P}W_{1})W_{2} 
\right) + exp\left( \sigma(\mathbf{P^{aug}}W_{1})W_{2}  \right)}.
\label{t-alpha}
\end{equation}

By the way, the $\mathbf{P}$ used in LCC in equation \ref{eq.scc} to generate $\mathbf{L^{lcc}}$ ($\mathbf{P^{lcc}}$) comes from equation \ref{eq.parallel}, which is same as the traditional self-train method \cite{pseudo-label}. And in order to avoid confusion between the two paradigms of GBC and GNN we do not use the $\mathbf{P^{fuse}}$ to generate the $\mathbf{P^{lcc}}$.

In backward propagation, there are three losses that supervising the neural network, they are vanilla loss $Loss_{train}$, anchor loss $Loss_{anchor}$ and the lcc psuedo label loss $Loss_{scc}$. 

\begin{equation}
    \begin{cases}
     loss_{lcc} =  CrossEntropy(\mathbf{P^{fuse}, L^{lcc}});\\
     loss_{anchor}=CrossEntropy(\mathbf{P^{anchor}, L^{anchor}});\\
     loss_{train} =CrossEntropy(\mathbf{P, L^{train}}).
    \end{cases}
    \label{AR}
\end{equation}

Specially, the $\mathbf{P^{anchor}}$ is the subset of $\mathbf{P^{aug}}$ which is the anchor part exclude the vanilla samples, and the $\mathbf{L^{lcc}}$ is additional supervisions beyond vanilla train label i.e., $\mathbf{L^{lcc}} \cap \mathbf{L^{train}} = \emptyset $.

The total loss is weighted sum by these three parts through the hyper-parameter $\beta$ and $\gamma$, it can be formulated as fallow:
\begin{equation}
    loss = loss_{train}+\beta\cdot loss_{anchor}+\gamma\cdot loss_{lcc}.
\end{equation}

\subsection{Complexity analysis}

In this section, we analyze the computational complexity of SCGNN from three aspects: 
(1) time complexity of the preprocessing;
(2) time complexity of the model training;
(3) space complexity.

\subsubsection{Preprocessing complexity}

Traditional semantic consistency mining methods, such as $k$-nearest neighbors ($k$NN) and other full search algorithms (FSA), rely on exhaustive pairwise similarity computation. Given $n$ nodes with feature dimension $d$, the similarity matrix construction requires: $\mathcal{O}(n^2 d).$ In addition, neighbor selection (e.g., top-$k$ selection) introduces extra sorting or selection overhead, which is typically $\mathcal{O}(n k \log n)$. Therefore, the overall complexity of FSA-based methods is dominated by:
\begin{equation}
\label{complexity.fsa}
    \mathcal{O}(n^2 d).
\end{equation}
This quadratic complexity severely limits scalability on large-scale graphs.

In contrast, the proposed GBC adopts an iterative partitioning strategy instead of global pairwise comparison. 

Specifically, GBC recursively splits granular balls using local clustering (e.g., $k$-means). For a granular ball of size $m$, the cost of one split is: $\mathcal{O}(m d k t),$ where $k$ is the number of clusters and $t$ is the number of iterations.

Since GBC partitions the dataset into multiple granular balls, the total cost can be expressed as:$\sum_i \mathcal{O}(m_i d k t),$ where$\sum_i m_i = n.$
Therefore, the overall complexity can be approximated as:
$\mathcal{O}(n d k t).$

In addition, the complexity of anchor-based augment graph mainly consists of three parts:

\textbf{(1) Anchor construction.} Each granular ball generates one anchor by averaging node features, which requires a single pass over all nodes: $\mathcal{O}(n d);$ \textbf{(2) Projection edges.} Each node is connected to its corresponding anchor, resulting in: $\mathcal{O}(n);$ \textbf{(3) Bridging edges.} Anchors belonging to the same class are connected, in the worst case (full connection), the number of edges is: $\mathcal{O}(B^2),$ where $B$ is the number of anchors, and $B = \mathcal{O}(\sqrt{n})$, this yields: $\mathcal{O}(n).$ Overall, the structure enhancement complexity is: $\mathcal{O}(n d).$

Therefore, the overall complexity of GBC method (include the structure enhancement) is:
\begin{equation}
\label{complexity.gbc}
    \mathcal{O}(n d k t),
\end{equation}
which is near-linear with respect to the number of nodes.

\textbf{Comparison.} Compared with FSA-based methods (Eq. \ref{complexity.fsa}), GBC reduces the preprocessing complexity from quadratic to near-linear (Eq. \ref{complexity.gbc}), providing a significantly more scalable solution for semantic consistency extraction. 

\subsubsection{Training complexity}

During training, SCGNN performs parallel forward propagation on both the vanilla graph $\mathbf{G}$ and the augment graph $\mathbf{G_{aug}}$.

For standard message-passing GNNs, the computational complexity is: $\mathcal{O}(|E| d).$ For the augment graph, the cost of one forward pass on $G_{aug}$ is: $\mathcal{O}((|E| + n)d).$ Since SCGNN processes both graphs, the total training complexity is: $\mathcal{O}(2|E|d + n d),$ which can be simplified to: 
\begin{equation}
    \mathcal{O}(|E| d).
\end{equation}

Additional components, such as label consistency checking (LCC) and prediction fusion, introduce only linear overhead $\mathcal{O}(n)$ or $\mathcal{O}(n d)$, which are negligible compared to GNN computation.





\subsubsection{Space complexity.}
In addition to time complexity, SCGNN also exhibits favorable space complexity. 
Unlike FSA-based methods that require $\mathcal{O}(n^2)$ memory to store pairwise similarity matrices, GBC operates without constructing such dense structures and only requires $\mathcal{O}(n d)$ memory. This further explains the superior memory efficiency observed in Table \ref{tab.scale}.

The anchor-based augment graph introduces only $\mathcal{O}(n)$ additional nodes and edges, resulting in an overall space complexity comparable to standard GNNs.

\section{Experiments}
\subsection{Setup}
\label{sec.setup}
In this section, we will introduce some basic setup of next experiments, train settings, dataset, backbone and baseline will be introduce in turn.

\textbf{Setting.} In all experiments, we chose the adam \cite{adam} as optimizer, fixed the max train epochs to $200$, fixed the hidden neuro number of backbone to $64$, learn rate fixed into $0.01$, weight decay fixed into $5\times 10^{-4}$, dropout fixed into $0.5$, and in reasoning stage we chose the best performing epoch on the validation set. All experiments were conducted on a machine with 90 GB RAM and an NVIDIA A100 GPU with 40 GB memory.

\textbf{Dataset.} We take 11 datasets in total, including: Cora, Citeseer, Pubmed \cite{planetoid}, Wiki \cite{AttributedGraph}, Physics \cite{coauthor}, Cornell, Wisconsin, Texas \cite{geomgcn_webkb}, ogbn-arxiv \cite{arxiv13, arxiv20}, Roman-empire and Amazon-ratings \cite{heterophilous}.  Cora, CiteSeer, PubMed, Cornell, Wisconsin, Texas, Roman-empire, Amazon-ratings and ogbn-arxiv fallow the default train splits, others without default splits we adopt $20\%$ for train, $40\%$ for validate and $40\%$ for test. The more detailed dataset descriptions are organized in Table \ref{dataset}.

\begin{table*}
\caption{The details of datasets used in this paper.}
\begin{tabular*}{\tblwidth}{LLLLLL}
\toprule
Dataset&Nodes&Edges&Features&Class&label rate \\
\midrule
Cora&2708&10556&1433&7&5\%\\
CiteSeer&3327&9104&3703&6&4\%\\
PubMed&19717&88648&500&3&0.3\%\\
Wiki&2405&17981&4973&17&20\%\\
Physics&34493&495924&8415&5&20\%\\
ogbn-arxiv&169343&1166243&128&40&54\%\\
Cornell&183&298&1703&5&48\%\\
Wisconsin&251&515&1703&5&48\%\\
Texas&183&325&1703&5&48\%\\
Roman-empire&22662&65854&300&18&50\%\\
Amazon-ratings&24492&186100&300&5&50\%\\
\bottomrule
\end{tabular*}
\label{dataset}
\end{table*}

\textbf{Backbone.} We take many classic GNN models as the backbone of our SCGNN to validate the efficiency , the GNNs includes: GCN \cite{gcn}, MLP \cite{MLP,MLP2}, GAT \cite{gat}, APPNP \cite{appnp}, GIN \cite{gin}, GraphSAGE \cite{graphsage} and H$_2$GCN \cite{h2gcn}.

\textbf{Baseline.} All evaluations may be concern with three baseline, and all baselines compared with are all plug-in model as same as ours, including: BORF \cite{BORF}, ComFy \cite{comfy} and AGMixup \cite{agmixup}. All of these baseline methods focus on data augmentation. Specifically, BORF \cite{BORF} introduced a Batch Ollivier-Ricci Flow-based rewiring algorithm that simultaneously mitigates both over-smoothing and over-squashing issues, thereby enhancing the expressive capability of GNNs. ComFy \cite{comfy} proposed a hybrid rewiring strategy that leverages both community structure and feature similarity, aiming to strengthen local feature coherence while preserving the global community structure to improve GNN performance. AGMixup \cite{agmixup} presented an adaptive graph mixup framework that employs data mixing techniques to augment graph representations and further boost the generalization ability of GNNs.

\subsection{Accuracy evaluation}
First of all, we give the numerical evaluations about the accuracy of semi-supervised node classification task, concerned with 8 datasets, the results are taken as average of 10 runs. The numerical statistics can be seen in Table \ref{classification}. Compared with the SOTA models, our model achieves consistent superiority in accuracy.

\begin{table*}
\caption{The semi-supervised classification accuracy (\%) of SCGNN and other all baselines with all seven GNN backbones across all datasets. \fbox{Frame} is the overall best across all settings, \textbf{bold} is the best in same backbone, \underline{underline} is the second-best in same backbone, and variance is written at the right bottom corner.}
\begin{tabular*}{\tblwidth}{LLLLLLLLLLL}
\toprule
Backbone&Method&Cora&CiteSeer&PubMed&Wiki&Cornell&Wisconsin&Texas&Roman-empire\\

\midrule
\multirow{4}{*}{MLP}&BORF &54.9$_{\pm0.82}$&52.31$_{\pm0.44}$&70.38$_{\pm0.25}$&78.48$_{\pm0.35}$&
76.21$_{\pm1.30}$&80.39$_{\pm0.37}$&78.38$_{\pm0.00}$&64.57$_{\pm0.14}$\\
&ComFy &\underline{57.20}$_{\pm0.55}$&\underline{58.96}$_{\pm0.65}$&\underline{72.64}$_{\pm0.20}$&\underline{78.86}$_{\pm0.37}$&
\underline{77.84}$_{\pm4.16}$&\underline{81.18}$_{\pm1.40}$&\underline{80.00}$_{\pm1.93}$&\underline{66.03}$_{\pm0.16}$\\
&AGMixup &56.67$_{\pm0.17}$&53.03$_{\pm0.66}$&70.97$_{\pm0.90}$&73.63$_{\pm0.91}$&
73.87$_{\pm4.59}$&79.08$_{\pm0.92}$&74.77$_{\pm2.55}$&65.59$_{\pm0.05}$\\
&SCGNN &\textbf{67.11}$_{\pm4.69}$&\textbf{63.65}$_{\pm4.51}$&\textbf{75.66}$_{\pm0.45}$&\fbox{\textbf{79.11}}$_{\pm0.41}$&
\fbox{\textbf{78.38}}$_{\pm3.60}$&\fbox{\textbf{82.35}}$_{\pm4.00}$&\fbox{\textbf{83.78}}$_{\pm4.09}$&\textbf{66.33}$_{\pm0.00}$\\

\midrule
\multirow{4}{*}{GCN}&BORF &79.07$_{\pm0.20}$&65.2$_{\pm0.42}$&76.89$_{\pm0.15}$&\underline{73.76}$_{\pm0.36}$&
38.91$_{\pm1.6}$&43.92$_{\pm1.7}$&39.46$_{\pm1.4}$&46.04$_{\pm0.46}$\\
&ComFy &80.16$_{\pm0.66}$&\underline{69.42}$_{\pm0.51}$&\underline{78.06}$_{\pm0.39}$&72.87$_{\pm0.53}$&
\underline{48.11}$_{\pm4.93}$&\underline{59.61}$_{\pm1.52}$&\underline{62.70}$_{\pm2.37}$&\underline{49.06}$_{\pm1.76}$\\
&AGMixup &\underline{81.42}$_{\pm0.60}$&68.26$_{\pm1.00}$&77.80$_{\pm0.35}$&66.44$_{\pm1.30}$&
40.54$_{\pm0.00}$&52.55$_{\pm2.29}$&60.00$_{\pm4.32}$&43.96$_{\pm0.19}$\\
&SCGNN &\fbox{\textbf{84.00}}$_{\pm0.21}$&\textbf{71.23}$_{\pm2.02}$&\textbf{79.71}$_{\pm0.12}$&\textbf{76.61}$_{\pm0.17}$&
\textbf{67.57}$_{\pm1.93}$&\textbf{66.67}$_{\pm1.77}$&\textbf{70.27}$_{\pm2.92}$&\textbf{54.43}$_{\pm0.11}$\\

\midrule
\multirow{4}{*}{GAT}&BORF &76.54$_{\pm0.68}$&66.46$_{\pm0.70}$&75.77$_{\pm0.43}$&\underline{68.45}$_{\pm0.61}$&
\underline{45.95}$_{\pm2.02}$&43.92$_{\pm2.43}$&40.54$_{\pm3.89}$&52.12$_{\pm0.70}$\\
&ComFy &73.97$_{\pm1.89}$&66.46$_{\pm1.71}$&75.60$_{\pm1.13}$&56.67$_{\pm5.40}$&
43.78$_{\pm3.35}$&49.41$_{\pm2.05}$&52.97$_{\pm3.28}$&\underline{58.38}$_{\pm0.38}$\\
&AGMixup &\underline{79.63}$_{\pm0.91}$&\underline{68.70}$_{\pm0.24}$&\underline{77.17}$_{\pm1.10}$&63.58$_{\pm2.82}$&
43.78$_{\pm6.71}$&\underline{52.55}$_{\pm7.27}$&\underline{61.62}$_{\pm6.92}$&58.23$_{\pm0.78}$\\
&SCGNN &\textbf{82.61}$_{\pm1.99}$&\textbf{70.40}$_{\pm3.77}$&\textbf{78.19}$_{\pm0.26}$&\textbf{71.83}$_{\pm1.02}$&
\textbf{64.86}$_{\pm2.21}$&\textbf{64.71}$_{\pm1.36}$&\textbf{70.27}$_{\pm4.02}$&\textbf{65.85}$_{\pm0.02}$\\

\midrule
\multirow{4}{*}{GIN}&BORF &\underline{75.91}$_{\pm0.87}$&63.18$_{\pm1.16}$&75.45$_{\pm1.46}$&\underline{62.33}$_{\pm0.30}$&
41.89$_{\pm1.15}$&38.82$_{\pm1.45}$&38.65$_{\pm1.34}$&51.45$_{\pm0.25}$\\
&ComFy &74.92$_{\pm0.75}$&\underline{63.94}$_{\pm0.93}$&\underline{76.86}$_{\pm0.63}$&23.51$_{\pm2.39}$&
\underline{51.89}$_{\pm1.81}$&\underline{58.04}$_{\pm4.52}$&\underline{65.95}$_{\pm2.47}$&\underline{62.94}$_{\pm3.52}$\\
&AGMixup &73.40$_{\pm0.29}$&60.23$_{\pm1.07}$&75.93$_{\pm0.66}$&23.70$_{\pm1.47}$&
48.65$_{\pm2.21}$&54.25$_{\pm5.15}$&63.06$_{\pm2.55}$&58.41$_{\pm0.71}$\\
&SCGNN
&\textbf{78.48}$_{\pm3.64}$&\textbf{68.15}$_{\pm1.80}$&\textbf{78.91}$_{\pm0.62}$&\textbf{63.2}$_{\pm8.42}$&
\textbf{70.27}$_{\pm7.01}$&\textbf{72.55}$_{\pm1.54}$&\textbf{72.97}$_{\pm2.26}$&\textbf{63.54}$_{\pm8.31}$\\

\midrule
\multirow{4}{*}{APPNP}&BORF &79.57$_{\pm0.15}$&66.27$_{\pm0.27}$&76.17$_{\pm0.17}$&\underline{75.81}$_{\pm0.23}$&
38.65$_{\pm0.78}$&47.84$_{\pm1.26}$&41.08$_{\pm1.03}$&58.56$_{\pm0.53}$\\
&ComFy &80.00$_{\pm0.17}$&69.62$_{\pm0.52}$&\underline{78.04}$_{\pm0.38}$&74.01$_{\pm0.55}$&
\underline{54.05}$_{\pm3.06}$&\underline{63.92}$_{\pm1.79}$&58.38$_{\pm1.18}$&\underline{58.95}$_{\pm0.43}$\\
&AGMixup &\underline{81.87}$_{\pm0.62}$&\underline{69.87}$_{\pm1.14}$&77.67$_{\pm0.52}$&69.89$_{\pm1.99}$&
39.64$_{\pm1.27}$&52.94$_{\pm4.24}$&\underline{60.36}$_{\pm2.55}$&54.42$_{\pm0.05}$\\
&SCGNN &\textbf{83.22}$_{\pm1.72}$&\fbox{\textbf{71.91}}$_{\pm3.59}$&\fbox{\textbf{79.74}}$_{\pm0.11}$&\textbf{76.51}$_{\pm0.24}$&
\textbf{67.57}$_{\pm8.77}$&\textbf{68.63}$_{\pm4.00}$&\textbf{70.27}$_{\pm4.21}$&\textbf{59.24}$_{\pm0.04}$\\

\midrule
\multirow{4}{*}{SAGE}&BORF &77.35$_{\pm0.35}$&65.61$_{\pm0.72}$&73.72$_{\pm0.56}$&\underline{77.48}$_{\pm0.14}$&
69.46$_{\pm1.72}$&71.96$_{\pm1.57}$&72.97$_{\pm2.02}$&74.44$_{\pm0.28}$\\
&ComFy &78.02$_{\pm0.83}$&67.12$_{\pm0.71}$&76.76$_{\pm0.67}$&76.20$_{\pm0.79}$&
\underline{75.14}$_{\pm4.69}$&\underline{77.65}$_{\pm2.38}$&\underline{81.49}$_{\pm3.06}$&\underline{76.41}$_{\pm0.32}$\\
&AGMixup &\underline{79.03}$_{\pm0.09}$&\underline{68.13}$_{\pm0.93}$&\underline{77.37}$_{\pm0.42}$&67.81$_{\pm4.43}$&
65.77$_{\pm5.55}$&77.12$_{\pm1.85}$&78.38$_{\pm2.21}$&76.29$_{\pm0.29}$\\
&SCGNN &\textbf{79.73}$_{\pm2.53}$&\textbf{68.20}$_{\pm2.69}$&\textbf{77.62}$_{\pm0.42}$&\textbf{78.59}$_{\pm0.28}$&
\fbox{\textbf{78.38}}$_{\pm4.67}$&\textbf{78.43}$_{\pm1.13}$&\fbox{\textbf{83.78}}$_{\pm7.60}$&\textbf{76.80}$_{\pm0.18}$\\

\midrule
\multirow{4}{*}{H$_2$GCN}&BORF &76.55$_{\pm0.43}$&63.72$_{\pm0.42}$&\underline{77.69}$_{\pm0.58}$&70.08$_{\pm0.80}$&
\underline{67.57}$_{\pm1.53}$&\underline{69.61}$_{\pm0.62}$&77.57$_{\pm0.78}$&76.04$_{\pm0.18}$\\
&ComFy &76.26$_{\pm0.77}$&\underline{66.80}$_{\pm1.52}$&75.86$_{\pm1.20}$&48.13$_{\pm1.97}$&
61.08$_{\pm3.93}$&67.84$_{\pm1.40}$&\underline{77.62}$_{\pm2.82}$&\underline{76.56}$_{\pm0.35}$\\
&AGMixup &\underline{78.07}$_{\pm0.31}$&63.23$_{\pm0.80}$&76.40$_{\pm0.29}$&\underline{71.28}$_{\pm0.34}$&
66.67$_{\pm1.27}$&64.05$_{\pm4.03}$&77.48$_{\pm3.37}$&76.51$_{\pm0.22}$\\
&SCGNN &\textbf{78.10}$_{\pm0.29}$&\textbf{67.31}$_{\pm0.68}$&\textbf{77.91}$_{\pm2.17}$&\textbf{71.41}$_{\pm2.27}$&
\textbf{72.97}$_{\pm2.92}$&\textbf{74.51}$_{\pm0.62}$&\textbf{78.38}$_{\pm2.13}$&\fbox{\textbf{76.89}}$_{\pm5.68}$\\

\bottomrule
\end{tabular*}
\label{classification}
\end{table*}

\subsection{Scalability on large scale datasets}
To evaluate the scalability of SCGNN, we conduct experiments on several large-scale datasets, including Amazon-ratings, Physics and ogbn-arxiv. The results are reported in Table \ref{tab.large_scale}.

As shown in Table \ref{tab.large_scale}, SCGNN consistently improves the performance of the GCN backbone across all datasets where results are available, demonstrating its effectiveness on large-scale graphs. 



\begin{table}
\caption{Performance comparison between vanilla GCN and GCN enhanced with SCGNN on large-scale datasets.}
\begin{tabular*}{\tblwidth}{LLL}
\toprule
Dataset&GCN&SCGNN \\
\midrule
Amazon-ratings&42.93&\textbf{44.03}\\
Physics&95.98&\textbf{96.34}\\
ogbn-arxiv&52.01&\textbf{53.56}\\
\bottomrule
\end{tabular*}
\label{tab.large_scale}
\end{table}

\subsection{Ablation analysis}
To gain a deeper understanding of the contributions of different parts of the SCGNN, the ablation analysis will be focus on the structure enhancement, the supervision enhancement and the parallel strategy. The \textbf{Abla.1} is W/O the supervision enhancement (LCC labels); the \textbf{Abla.2} is W/O the structure enhancement (augment graph); the \textbf{Abla.3} is W/O the parallel strategy (vanilla graph channel). Results are shown in Table \ref{tab.ablation}.

\begin{table}
\caption{These ablation experiments are based on GCN backbone, \textbf{bold} is the best and \underline{underline} is the second best.}
\begin{tabular*}{\tblwidth}{LLLLL}
\toprule
Dataset&SCGNN&Abla.1&Abla.2&Abla.3 \\
\midrule
Cora&\textbf{84.00}&\underline{82.81}&81.65&82.13\\
CiteSeer&\textbf{71.23}&\underline{70.55}&70.12&69.10\\
PubMed&\textbf{79.71}&\underline{79.44}&79.40&78.82\\
Wiki&\textbf{76.61}&\underline{76.12}&75.05&76.09\\
Cornell&\textbf{67.57}&54.05&54.05&\underline{62.16}\\
Wisconsin&\textbf{66.67}&54.95&54.91&\underline{56.86}\\
Texas&\textbf{70.27}&\underline{67.57}&\underline{67.57}&\textbf{70.27}\\
Roman-empire&\textbf{53.38}&46.33&48.99&\underline{52.92}\\
\bottomrule
\end{tabular*}
\label{tab.ablation}
\end{table}

Removing either the supervision enhancement (Abla.1) or the structure enhancement (Abla.2) leads to consistent performance degradation across all datasets. This highlights the critical role of the proposed GBC-guided semantic consistency enhancement. Specifically, the structure enhancement injects group-level semantic consistency into the graph structure, while the supervision enhancement (LCC) provides reliable training signals to effectively exploit such semantic information. The noticeable performance drops in both cases indicate that semantic consistency must be both properly incorporated into the graph structure and effectively utilized during training.

In contrast, removing the parallel prediction strategy (Abla.3) results in relatively minor performance degradation, suggesting that this component mainly contributes to model robustness by balancing the vanilla and augment graph structures.

Overall, these results validate that the GBC-guided semantic consistency enhancement is the key factor in improving representation learning, while the parallel architecture further stabilizes the model.

\subsection{Label noise evaluation}
In this section, we evaluate the effectiveness of the proposed Label Consistency Check (LCC) from two perspectives: (1) the quality of pseudo-labels, measured by the noise rate, and (2) the resulting node classification performance. Specifically, we compare three types of predictions: the model prediction $\mathbf{P}$, the granular-ball-based prediction $\mathbf{P^{gbc}}$, and the LCC-refined prediction $\mathbf{P^{lcc}}$. The noise rate is defined as the ratio of incorrectly predicted labels to the total number of nodes. The results are reported in Table~\ref{tab.validation}.

\begin{table}
\caption{These validation experiments are based on GCN backbone. The indicators contains node classification accuracy and the label noise rate, and the $P^{lcc}$ means using $\mathbf{P^{lcc}}$ as additional label, the $P^{gbc}$ means using $\mathbf{P^{gbc}}$ as additional label, the $P$ means using $\mathbf{P}$ as additional label. \textbf{Blod} is the best.}
\begin{tabular*}{\tblwidth}{LLLLLLL}
\toprule
\multirow{2}{*}{Dataset}&\multicolumn{3}{L}{ACC$(\%)$}&\multicolumn3{L}{Noise$(\%)$}\\
\cmidrule(r){2-4}
\cmidrule(){5-7}
&$P^{lcc}$&$P^{gbc}$&$P$&$P^{lcc}$&$P^{gbc}$&$P$ \\
\midrule
Cora&\textbf{84.00}&82.25&83.81&\textbf{14}&49&21\\
CiteSeer&\textbf{71.23}&70.12&70.88&\textbf{36}&56&40\\
PubMed&\textbf{79.71}&79.21&79.68&\textbf{18}&31&24\\
Wiki&\textbf{76.61}&74.74&74.74&\textbf{8}&17&28\\
Cornell&\textbf{67.57}&64.86&40.54&\textbf{21}&26&42\\
Wisconsin&\textbf{66.67}&58.86&56.82&\textbf{20}&29&47\\
Texas&\textbf{70.27}&67.57&67.57&\textbf{27}&37&42\\
Roman-empire&\textbf{53.38}&52.43&48.75&\textbf{19}&32&61\\
\bottomrule
\end{tabular*}
\label{tab.validation}
\end{table}




According to the theoretical analysis in Proposition \ref{proposition}, 
LCC reduces the noise rate from a linear dependence on individual error rates to a quadratic dependence.
As shown in Table~\ref{tab.validation}, the noise rate of $\mathbf{P^{lcc}}$ is consistently lower than that of both $\mathbf{P}$ and $\mathbf{P^{gbc}}$ across all datasets. This demonstrates that LCC enhances pseudo-label quality by enforcing agreement between predictions and discarding inconsistent ones.

Furthermore, using $\mathbf{P^{lcc}}$ as additional supervision consistently leads to the best classification performance. In contrast, directly using $\mathbf{P}$ or $\mathbf{P^{gbc}}$ introduces higher noise, which negatively affects model training. These results indicate that, although $\mathbf{P}$ and $\mathbf{P^{gbc}}$ may individually contain considerable noise, their agreement subset provides a much more reliable supervision signal.

Overall, the experimental results strongly validate the property of LCC and confirm that it serves as an effective mechanism for generating high-quality pseudo-labels and improving model performance.

\subsection{Parameter sensitivity}
\begin{figure}
    \centering
    \includegraphics[width=0.9\columnwidth]{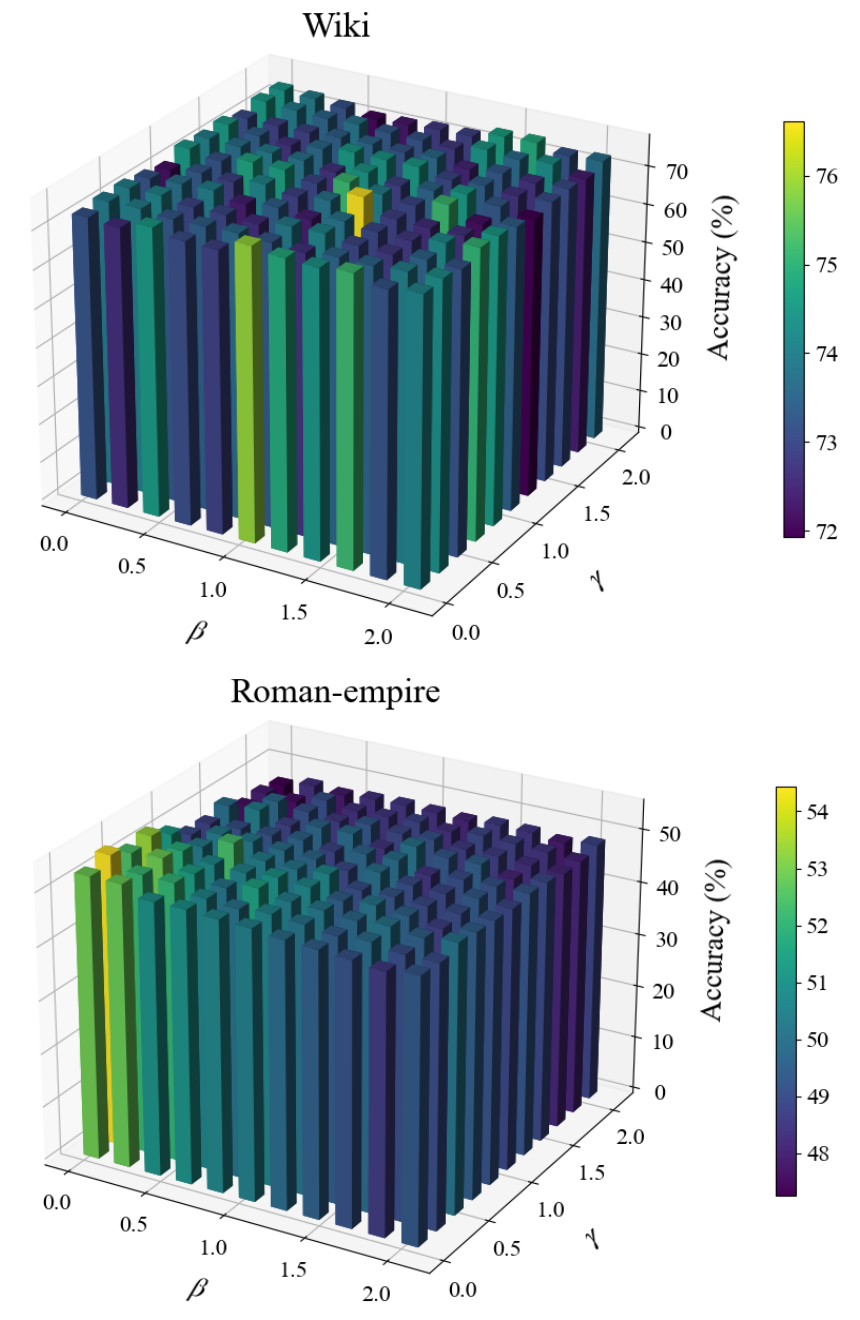}
    \caption{Top Figure is Wiki dataset and the bottom is the Roman-empire dataset.}
    \label{fig.para}
\end{figure}

Our model involves three hyper parameters they are purity  $p$ of $\mathbf{GBC(\cdot)}$ and weighted index of two losses $\beta$ and $\gamma$. The purity $p$ is be fixed in $0.8$, so in this section we mainly discuss the hyper parameter $\beta$ and $\gamma$. Results in shown in Figure \ref{fig.para}.

\section{Conclusion}
In this paper, we address semantic consistency mining in GNNs by introducing granular-ball computing (GBC) as a scalable alternative to $k$NN-based full search. By modeling group-level semantics, GBC reduces computational cost and improves robustness.

We further propose a plug-and-play framework, SCGNN, with a dual enhancement mechanism: structure enhancement via anchor-based graph structure construction and supervision enhancement via label consistency checking (LCC).

Extensive experiments demonstrate that SCGNN achieves superior performance with high efficiency. Ablation studies further verify the effectiveness of each component.

\section{Acknowledgement}
This work was supported by the National Science Foundation of China under Grant No.62466063, the Guizhou Provincial Department of Education Colleges and Universities Science and Technology Innovation Team (QJJ[2023]084).


\bibliographystyle{cas-model2-names}

\bibliography{cas-refs}



\end{document}